%% file: ms.tex
\documentclass[10pt,twocolumn,letterpaper]{article}

\usepackage{wacv}
\usepackage{times}
\usepackage{epsfig}
\usepackage{graphicx}
\usepackage{amsmath}
\usepackage{amssymb}
\usepackage[ruled]{algorithm2e}
\usepackage{amsmath,amssymb}
\usepackage{booktabs} 
\usepackage{subfigure}
\usepackage{arydshln}
\usepackage{multirow}

\wacvfinalcopy 

\ifwacvfinal
\def\assignedStartPage{1} 
\fi

\ifwacvfinal
\usepackage[breaklinks=true,bookmarks=false]{hyperref}
\else
\usepackage[pagebackref=true,breaklinks=true,colorlinks,bookmarks=false]{hyperref}
\fi

\ifwacvfinal
\else
\fi

\begin{document}

\title{REGroup: Rank-aggregating Ensemble of Generative Classifiers \\ for Robust Predictions}

\author{Lokender Tiwari$^{1,2}$
\and
Anish Madan$^{1}$
\and
Saket Anand$^{1}$
\and
Subhashis  Banerjee$^{3,4}$
\and
    {\tt\small $^{1}$IIIT-Delhi ~ $^{2}$TCS Research ~ $^{3}$IIT Delhi ~ $^{4}$Department of Computer Science, Ashoka University}
\and 
    {\tt\small \href{https://lokender.github.io/REGroup.html}{https://lokender.github.io/REGroup.html}} 
}

\maketitle
\input{sections/abstract}

\input{sections/introduction}

\input{sections/related_work}

\input{sections/method}

\input{sections/experiments}
\begin{appendix}

\input{sections/supp}

\end{appendix}

{\small
\bibliographystyle{ieee_fullname}
\bibliography{references}
}

\end{document}

%% file: sections/abstract.tex
\begin{abstract}
Deep Neural Networks (DNNs) are often criticized for being susceptible to adversarial attacks. Most successful defense strategies adopt adversarial training or random input transformations that typically require retraining or fine-tuning the model to achieve reasonable performance. In this work, our investigations of intermediate representations of a \emph{pre-trained} DNN lead to an interesting discovery pointing to intrinsic robustness to adversarial attacks. We find that we can learn a \emph{generative} classifier by statistically characterizing the neural response of an intermediate layer to clean training samples. The predictions of multiple such intermediate-layer based classifiers, when aggregated, show unexpected robustness to adversarial attacks. Specifically, we devise an ensemble of these generative classifiers that rank-aggregates their predictions via a \emph{Borda count}-based consensus. Our proposed approach uses a subset of the clean training data and a pre-trained model, and yet is agnostic to network architectures or the adversarial attack generation method. We show extensive experiments to establish that our defense strategy achieves state-of-the-art performance on the ImageNet validation set.
\end{abstract}

%% file: sections/introduction.tex
\section{Introduction}
\begin{figure}[!htb]
    \centering
    \includegraphics[width=\linewidth]{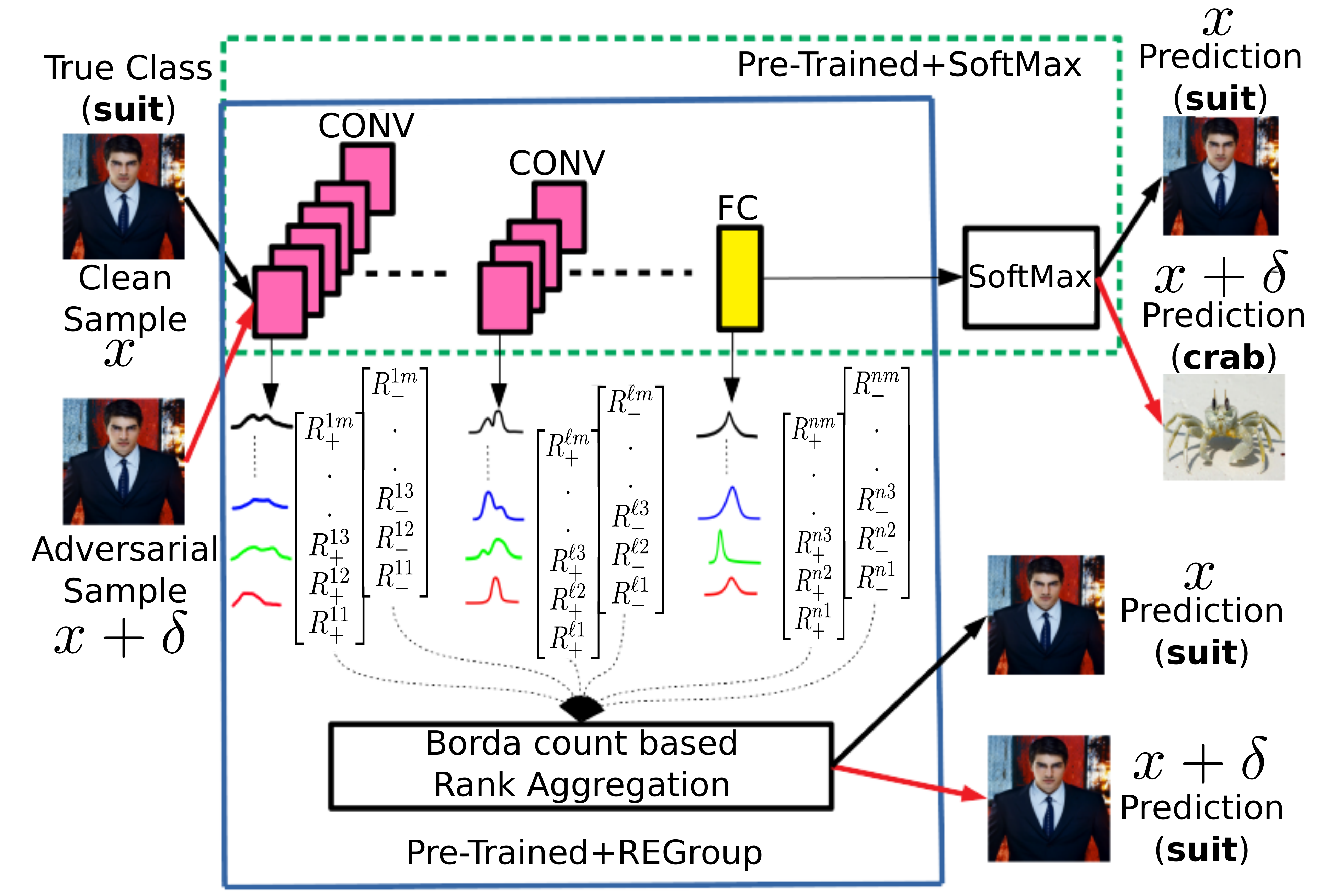}
    \caption{\textbf{Overview of REGroup.}~\textbf{R}ank-aggregating \textbf{E}nsemble of \textbf{G}enerative classifiers for \textbf{ro}b\textbf{u}st \textbf{p}redictions. REGroup uses a pre-trained network, and constructs layer-wise generative classifiers modeled by a mixture distribution of the positive and negative pre-activation neural responses at each layer. At test time, an input sample's neural responses are tested with generative classifiers to obtain ranking preferences of classes at each layer. These preferences are aggregated using \textit{Borda count} based preferential voting theory to make final prediction. \textit{Note:} construction of layer-wise generative classifiers is a one time process.
    }
    \label{fig:regroup}
\end{figure}

Deep Neural Networks (DNNs) have shown outstanding performance on many computer vision tasks such as image classification~\cite{krizhevsky2012imagenet}, speech recognition~\cite{hinton2012deep}, and video classification~\cite{karpathy2014large}. Despite showing superhuman capabilities in the image classification task~\cite{he2015delving}, the existence of \emph{adversarial examples} \cite{szegedy2013intriguing} have raised questions on the reliability of neural network solutions for safety-critical applications. 

Adversarial examples are carefully manipulated adaptations of an input, generated with the intent to fool a classifier into misclassifying them. One of the reasons for the attention that adversarial examples garnered is the ease with which they can be generated for a given model by simply maximizing the corresponding loss function. This is achieved by simply using a gradient based approach that finds a small perturbation at the input which leads to a large change in the output \cite{szegedy2013intriguing}. This apparent instability in neural networks is most pronounced for deep networks that have an accumulation effect over the layers. This effect results in taking the small, additive, adversarial noise at the input and amplifying it to substantially noisy feature maps at intermediate layers that eventually influences the softmax probabilities enough to misclassify the perturbed input sample. This observation of amplification of input noise over the layers is not new, and has been pointed out in the past \cite{szegedy2013intriguing}. The recent work by \cite{xie2019feature} addresses this issue by introducing \emph{feature denoising} blocks in a network and training them with adversarially generated examples. 

The iterative nature of generating adversarial examples makes their use in training to generate defenses computationally very intensive. For instance, the adversarially trained feature denoising model proposed by \cite{xie2019feature} takes 38 hours on 128 Nvidia V100 GPUs to train a baseline ResNet-101 with ImageNet. While we leverage this observation of noise amplification over the layers, our proposed approach \emph{avoids any training or fine-tuning} of the model. Instead, we use a representative subset of training samples and their layer-wise pre-activation responses to construct mixture density based generative classifiers, which are then combined in an ensemble using ranking preferences. 

Generative classifiers have achieved varying degrees of success as defense strategies against adversarial attacks. Recently, \cite{Fetaya_ICLR2020} studied the class-conditional generative classifiers and concluded that it is impossible to guarantee robustness of such models. More importantly, they highlight the challenges in training generative classifiers using maximum likelihood based objective and their limitations w.r.t. discriminative ability and identification of out-of-distribution samples. While we propose to use generative classifiers, we avoid using likelihood based measures for making classification decisions. Instead, we use rank-order preferences of these classifiers which are then combined using a \emph{Borda count}-based voting scheme. Borda counts have been used in collective decision making and are known to be robust to various manipulative attacks \cite{rothe2019borda}. 

In this paper, we present our defense against adversarial attacks on deep networks, referred to as \emph{Rank-aggregating Ensemble of Generative classifiers for robust predictions} (REGroup). At inference time, our defense requires white-box access to a pre-trained model to collect the pre-activation responses at intermediate layers to make the final prediction. We use the training data to build our generative classifier models. Nonetheless, our strategy is simple, network-agnostic, does not require any training or fine-tuning of the network, and works well for a variety of adversarial attacks, even with varying degress of hardness. Consistent with recent trends, we focus only on the ImageNet dataset to evaluate the robustness of our defense and report performance superior to defenses that rely on adversarial training \cite{kurakin2016adversarial} and random input transformation \cite{raff2019barrage} based approaches. Finally, we present extensive analysis of our defense with two different architectures (ResNet and VGG) on different targeted and untargeted attacks. Our primary contributions are summarized below:
\begin{itemize}
    \item We present REGroup, a retraining free, model-agnostic defense strategy that leverages an ensemble of generative classifiers over intermediate layers of the model.
    \item We model each layer-wise generative classifier as a simple mixture distribution of neural responses obtained from a subset of training samples. We discover that \emph{both positive and negative} pre-activation values contain information that can help correctly classify adversarially perturbed samples. 
    \item We leverage the robustness inherent in Borda-count based consensus over the generative classifiers.
    \item We show extensive comparisons and analysis on the ImageNet dataset spanning a variety of adversarial attacks.
\end{itemize}

%% file: sections/related_work.tex
\section{Related Work}
Several defense techniques have been proposed to make neural networks robust to adversarial attacks. Broadly, we can categorize them into two approaches that: 1. Modify training procedure or modify input before testing; 2. Modify network or change hyper-parameters and optimization procedure.


\subsection{Modify Training/Inputs During Testing} 

Some approaches of defenses in this category are mentioned below. \textit{Adversarial training}~\cite{zheng2016improving,moosavi2017universal,xie2020adversarial,shafahi2020universal}. \textit{Data compression}~\cite{bhagoji2018enhancing} suppresses the high-frequency components and presents an ensemble-based defense approach. \textit{Data randomization} ~\cite{wang2016learning,xie2017adversarial} based approaches apply random transformations to the input to defend against adversarial examples by reducing their effectiveness.

PixelDefend \cite{song2018pixeldefend} sets out to find the image with the highest probability within an $\epsilon$- neighbourhood of the original image, thereby moving the image back towards distribution seen in training data. Defense-GAN \cite{samangouei2018defense} tries to model the distribution of unperturbed images and at inference, it generates an image close to what was provided but without adversarial perturbations. These two methods use techniques to generate a clean version of the input and pass to the classifier. 
\subsection{Modify Network/Network Add-ons}
Defenses under this category address the \textit{detection} of adversarial attacks or cater to both \textit{detection and correction} of prediction.
The aim of detection only defenses is to highlight if an example is adversarial and prevent it from further processing. These approaches include employing a detector sub-network~\cite{metzen2017detecting}, training the main classifier with an outlier class~\cite{grosse2017statistical}, using convolution filter statistics~\cite{li2017adversarial}, or applying feature squeezing~\cite{xu2017feature} to detect adversarial examples. However, all of these methods have shown to be ineffective against strong adversarial attacks~\cite{carlini2017adversarial}\cite{sharma2018bypassing}.
Full defense approaches include applying defensive distillation~\cite{papernot2016distillation}\cite{papernot2017extending} to use the knowledge from the output of the network to re-train the original model and improve the resilience of a network to small perturbations. Another approach is to augment the network with a sub-network called Perturbation Rectifying Network (PRN) \cite{akhtar2018defense} to detect the perturbations; if the perturbation is detected, then PRN is used to classify the input image. However, later it was shown that the Carlini \& Wagner (C\&W) attack successfully defeated the defensive distillation approach.

\subsection{ImageNet Focused Defense Approaches}
A few approaches have been evaluated on the ImageNet dataset, most of which are based on input transformations or image denoising. Nearly all these defenses designed for ImageNet have failed a thorough evaluation, with a regularly updated list maintained at ~\cite{robustml}. The approaches in ~\cite{prakash2018deflecting} and \cite{liao2018defense} claimed 81\% and 75\% accuracy respectively under adversarial attacks. But after a thorough evaluation~\cite{athalye2018robustness} and accounting for obfuscated gradients~\cite{athalye2018obfuscated}, the accuracy for both was reduced to 0\%. Similarly, ~\cite{xie2017mitigating} and ~\cite{guo2017countering} claimed 86\% and 75\% respectively, but these were also reduced to 0\%~\cite{athalye2018obfuscated}. A different approach proposed in ~\cite{kannan2018adversarial} claimed an accuracy 27.9\% but later it was also reduced to 0.1\% ~\cite{engstrom2018evaluating}. For a comprehensive related work on attacks and defenses, we suggest reader to refer \cite{chakraborty2018adversarial}.

%% file: sections/method.tex
\newcommand{\mCpl}{\mathbb{C}^{+\ell}}
\newcommand{\mCnl}{\mathbb{C}^{-\ell}}

\newcommand{\mCpli}{\mathbb{C}^{+\ell i}}
\newcommand{\mCnli}{\mathbb{C}^{-\ell i}}

\newcommand{\bbR}{\mathbb{R}}
\newcommand{\mCp}{\mathcal{C}^{+}}
\newcommand{\mCn}{\mathcal{C}^{-}}
\newcommand{\mCpol}{\overline{\mathcal{C}}}
\newcommand{\mF}{\mathcal{F}}
\newcommand{\mFol}{\overline{\mathcal{F}}}
\newcommand{\bx}{x}
\newcommand{\tlbl}{$y$~}
\newcommand{\plbl}{$\widehat{y}$~}
\newcommand{\mL}{\boldsymbol{\phi}}
\newcommand{\mS}{\mathcal{S}}
\newcommand{\mT}{\mathcal{T}}
\newcommand{\mP}{\mathbb{P}}
\newcommand{\mN}{\mathbb{N}}
\newcommand{\bp}{\mbox{\mathbf{p}}}
\newcommand{\bn}{\mbox{\mathbf{n}}}
\newcommand{\mI}{\mathcal{I}}
\newcommand{\lth}{$\ell^{th}$~}
\newcommand{\ith}{$i^{th}$~}
\newcommand{\kth}{$k^{th}$~}
\newcommand{\jth}{$j^{th}$~}
\newcommand{\yth}{$y^{th}$~}
\newcommand{\dop}{$\mP^\ell$}
\newcommand{\don}{$\mN^\ell$}
\newcommand{\our}{REGroup}
\section{\our~ Methodology}

Well-trained deep neural networks have a hierarchical structure, where the early layers transform inputs to feature spaces capturing local or more generic patterns, while later layers aggregate this local information to learn more semantically relevant representations. In \our, we use many of the higher layers and learn class-conditional generative classifiers, which are simple mixture-distributions estimated from the pre-activation neural responses at each layer from a subset of training samples. An ensemble of these layer-wise generative classifiers is used to make the final prediction by performing a Borda count-based rank-aggregation. Ranking preferences have been used extensively in robust fitting problems in computer vision \cite{chin2011accelerated,Chin_NIPS2009,tiwari2018dgsac}, and we show its effectiveness in introducing robustness in DNNs against adversarial attacks. 

Fig. \ref{fig:regroup} illustrates the overall working of \our. The approach has three main components: First, we use each layer as a generative classifier that produces a ranking preference over all classes. Second, each of these class-conditional generative classifiers are modeled using a mixture-distribution over the neural responses of the corresponding layer. Finally, the individual layer's class ranking preferences are aggregated using Borda count-based scoring to make the final predictions. We introduce the notation below and discuss each of these steps in detail in the subsections that follow.

\noindent\textbf{Notation.} In this paper, we will {always} use $\ell$, $i$ and $j$ for indexing the \lth layer, \ith feature map and the \jth input sample respectively. The {true} and {predicted} class label will be denoted by \tlbl and \plbl respectively.

A classifier can be represented in a functional form as \plbl$=\mF(\bx)$, it takes an input $\bx$ and predicts its class label \plbl. We define $\mL^{\ell i}$ as the \lth layer's \ith {pre-activation feature map}, i.e., the neural responses \emph{before} they pass through the activation function. For convolutional layers, this feature map $\mL^{\ell i}$ is a 2D array, while for a fully connected layer, it is a scalar value.

\subsection{DNN Layers as Generative Classifiers}

\label{sec:LGC}
We use the highest $k$ layers of a DNN as generative classifiers that use the pre-activation neural responses to produce a ranking preference\footnote{A rank is assigned to each class based on a score. In the case of ImageNet dataset, the class with rank-1 is most preferred/likely class, while rank-1000 is the least preferred/likely class} over all classes. The layer-wise generative classifiers are modeled as a class-conditional mixture distribution, which is estimated using only a \textit{pre-trained} network and a small subset $\mS$ of the training data.

Let $\mS$ contain only correctly classified {training samples}\footnote{We took 50,000 out of $\sim$ 1.2 millions training images from ImageNet dataset, 50 per class.}, which we can further divide into $M$ subsets, one for each class i.e $\mS =\{  \cup_{y=1}^{M} \mS_{y}\}$, where $\mS_{y}$ is the subset containing samples that have labels $y$.

\subsubsection{Layerwise Neural Response Distributions}
Our preliminary observations indicated that while the ReLU activations truncate the negative pre-activations during the forward pass, these values still contain semantically meaningful information. Our ablative studies in Fig. \ref{fig:ablation} confirm this observation and additionally, on occasion, we find that the negative pre-activations are complementary to the positive ones. Since the pre-activation features are real-valued, we compute the features $\mL_j^{\ell i}$ for the \jth sample $x_j$, and define its positive ($P^{\ell i}_j$) and negative ($N^{\ell i}_j$) response accumulators as $P^{\ell i}_j = \sum\max(0,\mL_j^{\ell i})$, $N^{\ell i}_j=\sum\max(0,-\mL_j^{\ell i})$. 

For convolutional layers, these accumulators represent the overall strength of positive and negative pre-activation responses respectively, when aggregated over the spatial dimensions of the \ith feature map of the \lth layer. On the other hand, for the linear layers, the accumulation becomes trivial with each neuron having a scalar response $\mL_j^{\ell i}$. We can now represent the \lth layer by the positive and negative response accumulator vectors denoted by $P^\ell_j$ and $N^\ell_j$ respectively. We normalize these vectors and define the layer-wise probability mass function (PMF) for the positive and negative responses as $\mP^{\ell}_j=\frac{P^\ell_j}{||P^\ell_j||_1}$ and $\mN^{\ell}_j=\frac{N^\ell_j}{||N^\ell_j||_1}$ respectively. 

Our interpretation of $\mP^{\ell}_j$ and $\mN^{\ell}_j$ as a PMF could be justified by drawing an analogy to the softmax output, which is also interpreted as a PMF. However, it is worth emphasizing that we chose the linear rescaling of the accumulator vectors rather than directly applying a softmax normalization. By separating out the positive and negative accumulators, we obtain two independent representations for each layer, which is beneficial to our rank-aggregating ensemble discussed in the following sections. A softmax normalization over a feature map comprising of positive and negative responses would have entirely suppressed the negative responses, discarding all its constituent semantic information. An additional benefit of the linear scaling is its simple computation. Algorithm \ref{alg:LSS} summarizes the computation of the layer-wise PMFs for a given training sample.  

\begin{algorithm}[!htb]
\SetAlgoLined
\textbf{Input:} $\bx_j$ pre-activation features $\mL^{\ell i}_j\in\bbR^{H\times W}$\\
\For{ $\ell \in [1..n]$}{
 $P^{\ell i}_j = \sum\max(0,\mL_j^{\ell i}),~~~~~~\forall~ i$~~~ (sum over H, W) \\
 $N^{\ell i}_j=\sum\max(0,-\mL_j^{\ell i}),~~~~\forall~ i$~~ (sum over H, W) \\
 
}
$P^{\ell}_j \leftarrow P^{\ell}_j+\delta,$~~~~~$N^{\ell}_j \leftarrow N^{\ell}_j+\delta$ \\
$\mP_{j}^{\ell i} \leftarrow \frac{P^{\ell i}_j}{\sum_{i} \mP_{j}^{\ell i}},$~~~~~$\mN_{j}^{\ell i} \leftarrow \frac{N^{\ell i}_j}{\sum_{i} \mN_{j}^{\ell i}}$~~~(PMFs) 

\caption{Layerwise PMF of neural responses. $H\times W$ represents the spatial dimensions of pre-activation features. For \lth convolutional layer the dimensions of feature maps $H\times W = r^{\ell}\times s^{\ell}$, and for linear layers the dimensions of neuron output $H\times W = 1\times 1$. 
}
\label{alg:LSS}
\end{algorithm}

\subsubsection{Layerwise Generative Classifiers} 
We model the layerwise generative classifiers for class $y$ as a class-conditional mixture of distributions, with each mixture component as the PMFs $\mP^\ell_j$ and $\mN^\ell_j$ for a given training sample $x_j\in\mS_y$. The generative classifiers corresponding to the positive and negative neural responses are then defined as the following mixture of PMFs\\
\begin{equation}
    \mCpl_y = \sum_{j:\bx_j \in \mS_{y}}\lambda_j\mP_{j}^{\ell},\qquad
    \mCnl_y = \sum_{j:\bx_j \in \mS_{y}}\lambda_j\mN_{j}^{\ell}
\end{equation}
    
where the weights $\lambda_j$ are nonnegative and add up to one in the respective equations. Here, $\lambda_j$ is proportional to the softmax probability of the sample $x_j$, and  $\delta$ is the small constant used for numerical stability. We choose the weights to be proportional to the softmax probability value as predicted by the network given the input $x_j$. Using the subset of training samples $\mS$, we construct the class-conditional mixture distributions, $\mCpl_y$ and $\mCnl_y$ at each layer $\ell$ only once. At inference time, we input a test sample $x_j$, from the test set $\mT$, to the network and compute the PMFs $\mP^{\ell}_j$ and $\mN^{\ell}_j$ using Algorithm \ref{alg:LSS}. As our test input is a PMF and the generative classifier is also a mixture distribution, we simply use the KL-Divergence between the classifier model $\mCpl$ and the test sample $\mP^{\ell}_j$ as a classification score as
\begin{equation}
    P_{KL}(\ell,y) = \sum_{i} \mCpli_y \log \Bigg(\frac{\mCpli_y}{\mP^{\ell i}}\Bigg),\forall y\in\{1,\!\ldots,\!M\}
    \label{eq:klp}
\end{equation}
and similarly for the negative PMFs
\begin{equation}
    N_{KL}(\ell,y) = \sum_{i} \mCnli_y  \log \Bigg(\frac{\mCnli_y}{\mN^{\ell i}}\Bigg),\forall y \in\{1,\!\ldots,\!M\} 
    \label{eq:kln}
\end{equation}

We use a simple classification rule and select the predicted class \plbl as the one with the smallest KL-Divergence with the test sample PMF. However, rather than identifying \plbl, at this stage we are only interested in rank-ordering the classes, which we simply achieve by sorting the KL-Divergences (Eqns. (\ref{eq:klp}) and (\ref{eq:kln})) in ascending order. The resulting ranking preferences of classes for the \lth layer are given below in Eqns. \eqref{eq:rankvec_pos} and \eqref{eq:rankvec_neg} respectively. Where, $R^{\ell y}_{+}$ is the rank (position of $y^{th}$ class in the ascending order of KL-Divergences in $P_{KL}$) of $y^{th}$ class in the \lth layer preference list $R^{\ell}_{+}$. 
\begin{eqnarray}
R^{\ell}_{+}=[R^{\ell 1}_{+},R^{\ell 2}_{+},...,R^{\ell y}_{+},...,R^{\ell M}_{+}] \label{eq:rankvec_pos} \\
R^{\ell}_{-}=[R^{\ell 1}_{-},R^{\ell 2}_{-},...,R^{\ell y}_{-},...,R^{\ell M}_{-}]
\label{eq:rankvec_neg}
\end{eqnarray}

\subsection{Robust Predictions with Rank Aggregation}
Rank aggregation based {preferential voting} for making group decisions is widely used in selecting a winner in a democratic setup~\cite{rothe2019borda}. The basic premise of preferential voting is that $n$ voters are allowed to rank $m$ candidates in the order of their preferences. The rankings of all $n$ voters are then aggregated to make a final prediction.

{Borda count}~\cite{black1958theory} is one of the approaches for preferential voting that relies on aggregating the rankings of all the voters to make a collective decision~\cite{rothe2019borda,kahng2019statistical}. The other popular voting strategies to find a winner out of $m$ different choices include {Plurality voting}~\cite{van1992borda}, and {Condorcet winner}~\cite{young1988condorcet}. In Plurality voting, the winner would be the one who gets the maximum fraction of votes, while Condorcet winner is the one who gets the majority votes.

\subsubsection{Rank Aggregation Using Borda Count}
\textit{Borda count} is a generalization of the majority voting. In a two-candidates case it is equivalent to majority vote. The \textit{Borda count} for a candidate is the sum of the number of candidates ranked below it by each voter. In our setting, while processing a test sample $x_j\in\mT$, \textit{every layer} acts as two independent voters based on \dop and \don. The number of classes i.e $M$ is the number of candidates. The Borda count for the \yth class at the \lth layer is denoted by $B^{\ell y}=B^{\ell y}_{+}+B^{\ell y}_{-}$, where $B^{\ell y}_{+}$ and $B^{\ell y}_{-}$ are the individual Borda count of both the voters and computed as shown in eq. \eqref{eq:indv_bordacount}.
\begin{equation}
B^{\ell y}_{+}  = (M - R^{\ell y}_{+}),~~~~~~~
B^{\ell y}_{-}  = (M - R^{\ell y}_{-}) 
\label{eq:indv_bordacount}
\end{equation}

\subsubsection{Hyperparameter settings} 
We aggregate the Borda counts of highest $k$ layers of the network, which is the only hyperparameter to set in \our. Let $B^{:ky}$ denote the aggregated Borda count of \yth class from the last $k$ layers irrespective of the type (convolutional or fully connected). Here, $n$ is the total number of layers. The final prediction would be the class with maximum aggregated Borda count.
\begin{align}
\nonumber B^{:ky} & = \sum_{\ell=n-k+1}^{n} B^{\ell y}\\
\nonumber & = \sum_{\ell=n-k+1}^{n} B^{\ell y}_{+}+B^{\ell y}_{-},~~\forall y\in\{1..M\}\\
\widehat{y}&= \textit{argmax}_{y} ~~B^{:ky}
 \label{eq:agg_borda}
\end{align}
To determine the value of $k$, we evaluate \our~ on 10,000 \textit{correctly classified} samples from the ImageNet Validation set at {each layer}, using {per layer} Borda count i.e $ \widehat{y}= \arg\max_{y} ~~B^{\ell y}$. We select $k$ to be the number of later layers at which we get at-least $75\%$ accuracy. This can be viewed in the context of the confidence of individual layers on discriminating samples of different classes. We follow the above heuristic and found $k=5$ for both the architectures ResNet-50 and VGG-19, which we use in all our experiments. An ablation study with all possible values of $k$ is included in section \ref{sec:ablation}.

%% file: sections/experiments.tex
\newcommand{\ififtyk}{\text{V50K}}
\newcommand{\itenk}{\text{V10K}}
\newcommand{\itwok}{\text{V2K}}
\newcommand{\ionek}{\text{V1K}}

\section{Experiments}
\begin{figure}[!htb]
\centering
\begin{tabular}{c}
 \includegraphics[width=7.5cm]{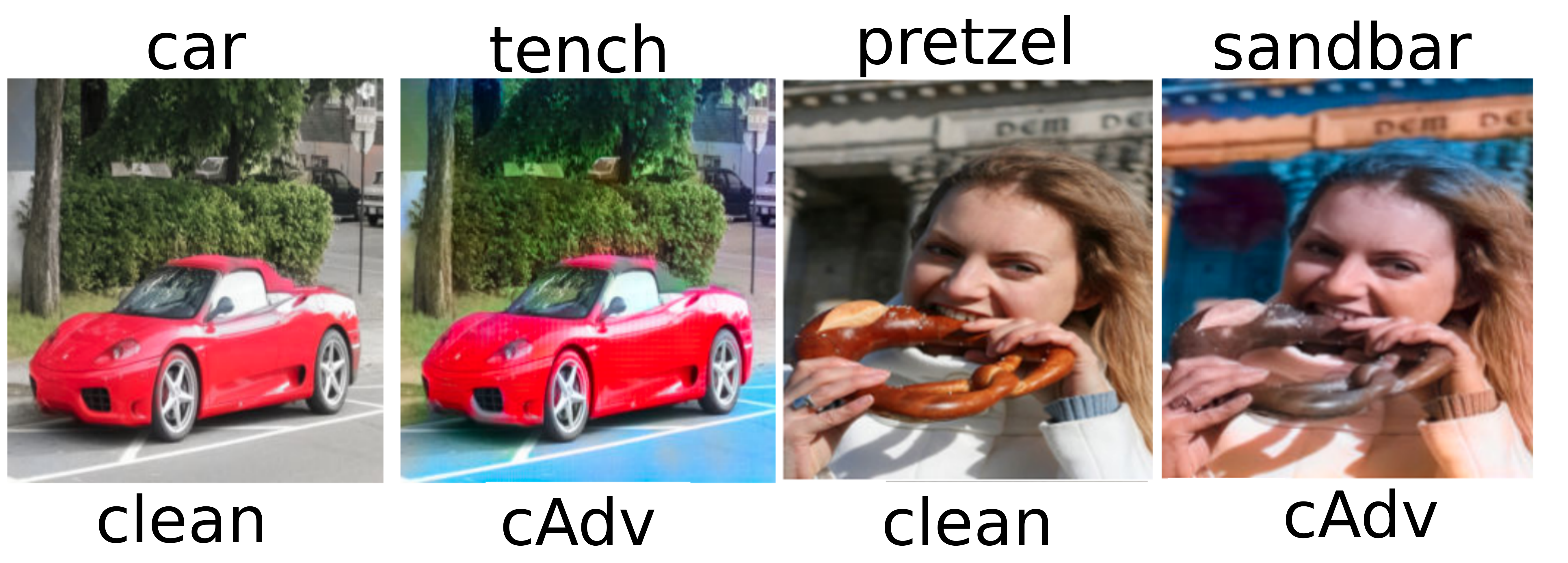} 
\end{tabular}
\caption{cAdv~\cite{bhattad2020unrestricted} adversarial examples}
\label{fig:cadv}
\end{figure}

In this section, we evaluate robustness of \our~ against state-of-the-art attack methods. We follow the recommendations on defense evaluation in~\cite{carlini2019evaluating}. \\
\noindent \textbf{Attack methods.} We consider attack methods in the following two categories: \textit{gradient-based} and \textit{gradient-free}. \\ 
\underline{{Gradient-Based Attacks}}. Within this category, we consider two variants, \textit{restricted} and \textit{unrestricted} attacks. The restricted attacks generate adversarial examples by searching an adversarial perturbations within the bound of $L_p$ norm, while unrestricted attacks generate adversarial example by manipulating image-based visual descriptors.  Due to restriction on the perturbation the adversarial examples generated by restricted attacks are similar to the clean original image, while unrestricted attacks generate natural-looking adversarial examples, which are far from the clean original image in terms of $L_p$ distance. We consider the following, \textit{Restricted attacks}: PGD~\cite{madry2017towards} , DeepFool~\cite{moosavi2016deepfool}, C\&W~\cite{carlini2017towards} and Trust Region~\cite{yao2019trust}, and \textit{Unrestricted attack}: cAdv~\cite{bhattad2020unrestricted} semantic manipulation attack. An example of cAdv is shown in Fig.~\ref{fig:cadv}. \\

\noindent \underline{{Gradient-Free Attacks}}. The approaches in this category does not have access to the network weights. We consider following attacks: SPSA~\cite{uesato2018adversarial}, Boundary~\cite{brendel2017decision} and Spatial~\cite{engstrom2019exploring}. Refer supplementary for the attack specific hyper-parameters detail.

\noindent \textbf{Network architectures.} We consider   ResNet-50\footnote{https://download.pytorch.org/models/resnet50-19c8e357.pth} and VGG-19\footnote{https://download.pytorch.org/models/vgg19-dcbb9e9d.pth} architectures, {pre-trained} on ImageNet dataset.\\
\noindent\textbf{Datasets.} We present our evaluations, comparisons and analysis only on ImageNet~\cite{imagenet} dataset. We use the subsets of full ImageNet validation set as described in Tab. \ref{tbl:dset}. Note:  V10K, V2K and V10C would be different for ResNet-50 and VGG-19, since an image classified correctly by ResNet-50 need not be classified correctly by the VGG-19. 
\begin{table}[h]
\centering
\resizebox{\linewidth}{!}{%
\begin{tabular}{l|l}
\hline
Dataset & Description            \\
\hline
V50K         & Full ImageNet validation set with 50000 images.                         \\
V10K         & A subset of 10000 correctly classified images from V50K set. 10 Per class. \\
V2K          & A subset of 2000 correctly classified images from V50K set. 2 Per class.\\
V10C         & A subset of correctly classified images of 10 sufficiently different classes.\\
\hline
\end{tabular}%
}
\caption{Dataset used for evaluation and analysis.}
\label{tbl:dset}
\end{table}

\subsection{Performance on Gradient-Based Attacks}

\noindent \textbf{Comparison with adversarial-training/ fine-tuning.} 
We evaluate \our~on clean samples as well as adversarial examples generated using PGD $(\epsilon = 16)$ from V50K dataset, and compare it with prior state-of-the-art works. The results are reported in Tab. \ref{tbl:sota}, and we see that \our~outperforms the state-of-the-art input transformation based defense BaRT~\cite{raff2019barrage}, both in terms of the clean and adversarial samples (except in the case of Top-1 accuracy with $\hat{k}=10$, which is the number of input transformations used in BaRT). We see that while our performance on clean samples decreases when compared to adversarial training (Inception v3), it improves significantly on adversarial examples with a high $\epsilon=16$. While our method is not directly comparable with adversarially trained Inception v3 and ResNet-152, because the base models are different, a similar decrease in the accuracy over clean samples is reported in their paper. The trade-off between robustness and the standard accuracy has been studied in \cite{dohmatob2018limitations} and \cite{tsipras2018robustness}.

An important observation to make with this experiment is, if we set aside the base models of ResNets and compare Top-1 accuracies on clean samples of full ImageNet validation set, our method (\our) without any \textit{adv-training/fine-tuning} either outperforms or performs similar to the state-of-the-art \textit{adv-training/fine-tuning} based methods \cite{raff2019barrage, xie2019feature}.
\begin{table}[!htb]
\centering

\scriptsize
\begin{tabular}{lcccc}
\toprule 
     ({Dataset used:} \ififtyk).           &  \multicolumn{2}{c}{Clean Images} &\multicolumn{2}{c}{Attacked Images}
                \vspace{0.01cm} \\
                
                \cline{2-5}
                \vspace{0.02cm}
                    Model            & Top-1 & Top-5 & Top-1 & Top-5 \vspace{0.01cm}\\
                                \toprule
                                ResNet-50 & 76 & 93 & 0.0 & 0.0 \\
                                Inception v3 & 78 & 94 & 0.7 & 4.4 \\
                               ResNet-152 & 79 & 94 & - & - \\
                               \hline
                               Inception v3 w/Adv. Train & 78 & 94 & 1.5 & 5.5 \\
                                ResNet-152 w/Adv. Train & 63 & - & 45 & - \\
                                ResNet-152 w/Adv. Train w/ denoise & 66 & - & 49 & - \\
                                ResNet-50-BaRT, $\hat{k}=5$ & 65 & 85 & 16 & 51 \\
                        ResNet-50-BaRT, $\hat{k}=10$ & 65 & 85 & 36 & 57 \\
                                \hline   
                                ResNet-50-\our & 66 & 86 & {22} & {65}\\
                               \toprule
\end{tabular}%
\caption{\textbf{Comparison with adversarially trained and fine-tuned classification models.} Top-1 and Top-5 classification accuracy (\%) of adversarial trained (Inception V3 \cite{kurakin2016adversarial} and ResNet-152 ~\cite{xie2019feature}) and fine-tuned (ResNet-50 BaRT~\cite{raff2019barrage}) classification models. Clean Images are the non-attacked original images. The results are divided into three blocks, the top block include original networks, middle block include defense approaches based on adversarial re-training/fine-tuning of original networks, bottom block is our defense \textit{without re-training/fine-tuning}. Results of the competing methods are taken from their respective papers. `-' indicate the results were not provided in the respective papers. }
\label{tbl:sota}
\end{table}

\begin{figure}[!htb]
    \centering
    \includegraphics[width=8cm]{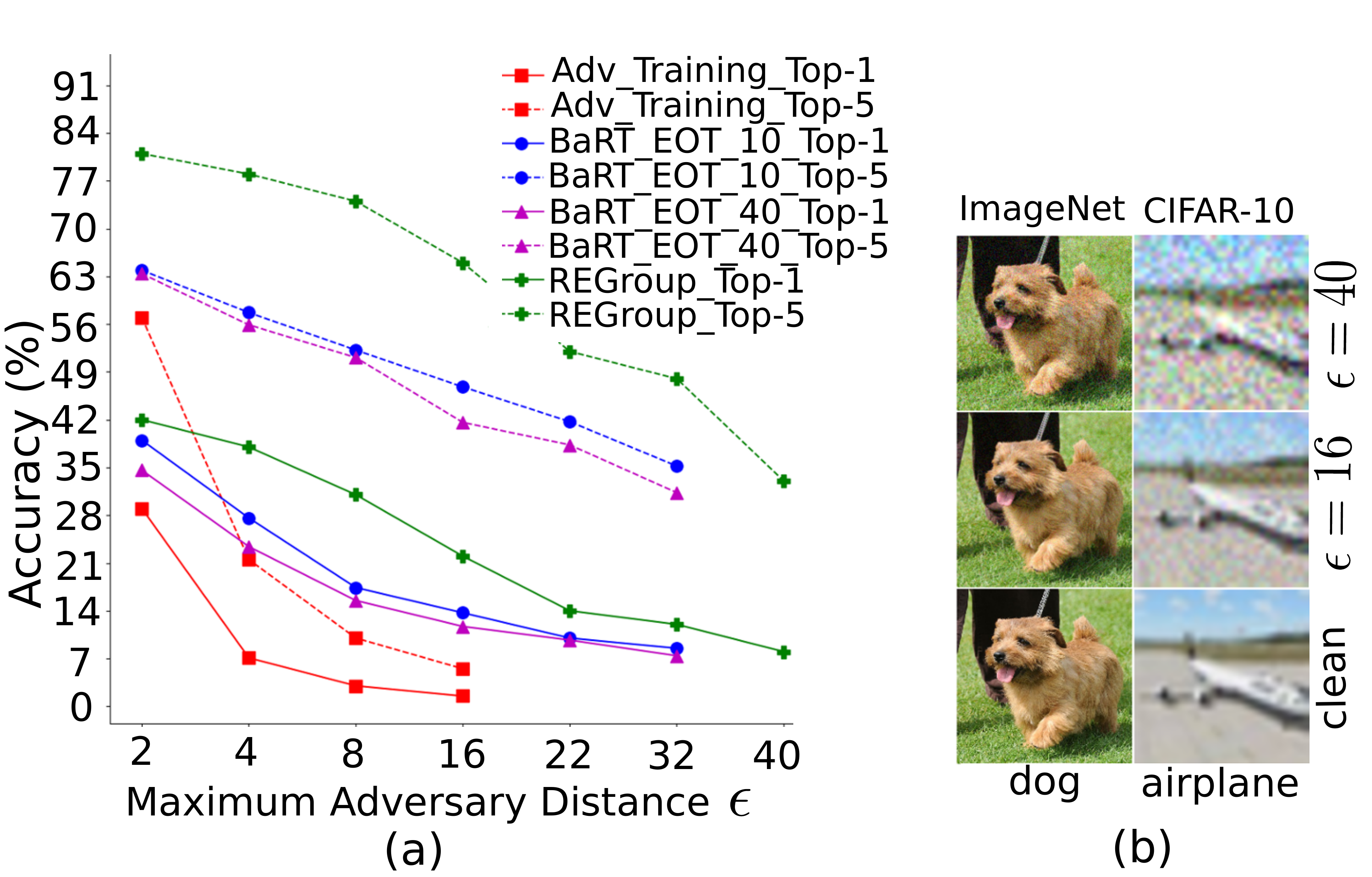}
    \caption{\textbf{Top-1 and Top-5 accuracy(\%) w.r.t PGD adversarial strength.} Comparison with adversarial training based method~\cite{kurakin2016adversarial} and fine-tuning using random input transformations based method (BaRT) \cite{raff2019barrage} with Expectation Over Transformation (EOT) steps 10 and 40, against the PGD perturbation strength ($\epsilon$). The results of the competing methods are taken from their respective papers. Dataset used: \ififtyk. }
    \label{fig:pgd_strength}
\end{figure}

\noindent \textbf{Performance w.r.t PGD Adversarial Strength.}
We evaluate \our~ w.r.t the maximum perturbation of the adversary. The results are reported in Fig.~\ref{fig:pgd_strength}(a).
\our~ outperforms both the adversarial training \cite{kurakin2016adversarial} and BaRT~\cite{raff2019barrage}.
Both adversarial training and BaRT have shown protection against PGD adversarial attacks with a maximum perturbation strength $\epsilon=16$ and $\epsilon=32$ respectively, however we additionally show the results with $\epsilon=40$ on full ImageNet validation set. 
We also note that with increasing perturbation strength, our defense's accuracy is also strictly decreasing. This is in accordance with \cite{carlini2019evaluating}, where transitioning from a clean image to noise should yield a downward slope in accuracy, else there could be some form of gradient masking involved. While it may seem $\epsilon=40$ is a large perturbation budget and it will destroy the object information in the image completely, but we would like to emphasize that it is not the case when using large size images. A comparison of PGD examples generated with $\epsilon=40$ using CIFAR-10 (32 $\times$ 32) and ImageNet (224 $\times$ 224) images is shown in Fig.~\ref{fig:pgd_strength}(b). \\

\begin{table}[h]
\centering
\setlength{\tabcolsep}{1.5pt}
\scriptsize
\begin{tabular}{l|ccc|ccc|ccc}
\hline
                  \multicolumn{1}{c}{}   &
                  \multicolumn{1}{c}{}   & \multicolumn{1}{c}{}   & \multicolumn{1}{c}{}   &  \multicolumn{3}{c|}{ResNet-50}                                                                                                                       & \multicolumn{3}{c}{VGG-19}                                                                                                                          \\

 \cline{5-10}   \multicolumn{1}{c}{}   &
 \multicolumn{1}{c}{}   &
 \multicolumn{1}{c}{UN /}   &
 \multicolumn{1}{c}{}   &  \multicolumn{1}{c}{}  & \multicolumn{1}{c}{SMax}                          & \multicolumn{1}{c|}{\our}                      & \multicolumn{1}{c}{}   &  \multicolumn{1}{c}{SMax}                          & \multicolumn{1}{c}{\our}                           \\
  \cline{5-10} \multicolumn{1}{c}{}   & 
     \multicolumn{1}{c}{Data}   & 
     \multicolumn{1}{c}{TA / HC}   & 
     \multicolumn{1}{c}{$\epsilon$}   & \multicolumn{1}{c}{\#S}  & \multicolumn{1}{c}{T1(\%)}                          & \multicolumn{1}{c|}{T1(\%)}                      & \multicolumn{1}{c}{\#S}   &  \multicolumn{1}{c}{T1(\%)}                          & \multicolumn{1}{c}{T1(\%)}                           \\
 \cline{1-2}
 \cline{3-10} 
Clean &V10K&-- & --  & 10000 &      {100}                          &     88                          &10000       &        {100}                           &        76
\\
Clean &V2K &--&-- &2000   & 100   &86 &2000                                  & 100       &      72                                    \\
Clean &V10C &--&-- &417   & 100   &84 &392                                  & 100       &      79                                    \\
\hdashline
PGD &V10K& UN &4 ($L_\infty$) &  9997   &  0                         & {48}                                              &9887         &        0                   &     {46}                                                                    \\
DFool   &V10K&UN & 2 ($L_2$)     &  9789     &       0                                   &     {61}                            &9939         &           0                            &        {55}                                        \\
C\&W &V10K &UN &4 ($L_2$)      & 10000   &       0                                &       {40}                             &10000       &      0                                 &    {38}                                                       \\
TR &V10K&UN &2 ($L_\infty$) & 10000    &      0                       &        {41}                              &  9103     &   0                             &       {45}                                     \\
cAdv &V10C&UN &-- & 417    &      0                       &        {37}                              &  392     &   0                             &       {18}                                     \\
\hdashline

PGD &V2K &TA & ($L_\infty$)&2000 &0       &47                 &2000                                           &0                           &       31                                                                 \\
C\&W  &V2K &TA &($L_2$) &2000 & 0     &46                    & 2000                       &0                                        &      38                                                   \\

\hdashline
PGD  &V2K &UN+HC& ($L_\infty$)&  2000   &  0                         & 21                                              &  2000       &        0                   &     19                                                                    \\
PGD  &V2K &TA+HC& ($L_\infty$) &  2000   &  0                         &   23                                            &  2000       &        0                   &   17                                                                      \\
\hline
\end{tabular}%
\caption{\textbf{Performance on Gradient-Based Attacks.} Comparison of Top-1 classification accuracy between SoftMax (SMax) and \our~ based final classification. UN and TA indicates, un-targeted and targeted attacks respectively.  The +HC indicates adversarial examples are generated with high-confidence ( $>90\%$) constraint, in this case $\epsilon$ can be any value that satisfies the HC criteria. For targeted attack we select a target class uniformly at random from the 1000 classes leaving out the true class. $\#S$ is the number of images for which the attacker is successfully able to generate adversarial examples using the respective attack models and the accuracies are reported with respect to the $\#S$ samples, hence the 0\% accuracies with the SoftMax (SMax). Since $\#S$ is different for several attacks, therefore, the performance may not be directly comparable \emph{across} different attacks. `--' indicate the information is not-applicable. For data description refer Tab. \ref{tbl:dset}.}
\label{tbl:grad-based}
\end{table}

\noindent \textbf{Performance on Un-Targeted Attacks.}
We evaluate \our~ on various untargeted attacks and report results in Tab. \ref{tbl:grad-based}. The perturbation budgets ($\epsilon$) and dataset used for the respective attacks are listed in the table. With the exception of the maximum perturbation allowed, we used default parameters given by FoolBox~\cite{rauber2017foolbox}. Due to space limitations, the attack specific hyper-parameters detail are included in the supplementary. We observe that the performance of our defense is quite similar for both the models employed. This is due to the attack-agnostic nature of our defense. We achieve 48\% accuracy (ResNet-50) for PGD attack using our defense which is significant given that PGD is considered to be one of the strongest attacks among the class of first order adversaries. \\

\noindent \textbf{Performance on Unrestricted, Untargeted Semantic Manipulation Attacks.}
We consider V10C dataset for cAdv attack. We use the publicly released source code by the authors. Specifically we use $\text{cAdv}_{4}$ variant with the parameters suggested by the authors. The results are reported in Tab. \ref{tbl:grad-based}. \\

\noindent \textbf{Performance on Targeted Attacks.}
We consider \itwok~dataset for targeted attacks and report the performance on PGD and C\&W targeted attacks in Tab. \ref{tbl:grad-based}. Target class for such attacks is chosen uniformly at random from the 1000 ImageNet classes apart from the original(ground-truth) class.  \\

\noindent \textbf{Performance on PGD attack with High Confidence.}
We evaluate \our~on PGD examples on which the network makes highly confident predictions using SoftMax. We generate un-targeted and targeted adversarial examples using PGD attack with a constraint that the network's confidence of the prediction of adversarial examples is \text{at-least 90}\%. For this experiment we do not put constraint on the adversarial perturbation i.e $\epsilon$. Results are reported in Tab. \ref{tbl:grad-based}.

\subsection{Performance on Gradient-Free Attacks}

Several studies \cite{athalye2018obfuscated}, \cite{papernot2017practical} have observed a phenomenon called\textit{ gradient masking}. This phenomenon occurs when a practitioner unintentionally or intentionally proposes a defense which does not have meaningful gradients, either by reducing them to small values (vanishing gradients), removing them completely (shattered gradients) or adding some noise to it (stochastic gradient). 

Gradient masking based defenses hinder the gradient computation and in turn inhibit gradient-based attacks, thus providing a false sense of security. Therefore, to establish the robustness of a defense against adversarial attacks in general, it is important to rule out that a defense relies on gradient masking. 

To ensure that \our~ is not masking the gradients we follow the standard practice \cite{pang2019rethinking} \cite{zhang2020clipped} and evaluate on strong gradient-free SPSA~\cite{uesato2018adversarial} attack. In addition to SPSA, we also show results on two more gradient-free attacks, Boundary~\cite{brendel2017decision} and Spatial~\cite{engstrom2019exploring} attack. The results are reported in Tab. \ref{tbl:grad-free}. 

\begin{table}[!htb]
\centering
\setlength{\tabcolsep}{1.1pt}
\scriptsize
\begin{tabular}{l|ccc|ccc|ccc}
\hline
                  \multicolumn{1}{c}{}   &
                  \multicolumn{1}{c}{}   & \multicolumn{1}{c}{}   & \multicolumn{1}{c}{}   &  \multicolumn{3}{c|}{ResNet-50}                                                                                                                       & \multicolumn{3}{c}{VGG-19}                                                                                                                          \\

 \cline{5-10}   \multicolumn{1}{c}{}   &
 \multicolumn{1}{c}{}   &
 \multicolumn{1}{c}{UN /}   &
 \multicolumn{1}{c}{}   &  \multicolumn{1}{c}{}  & \multicolumn{1}{c}{SMax}                          & \multicolumn{1}{c|}{\our}                      & \multicolumn{1}{c}{}   &  \multicolumn{1}{c}{SMax}                          & \multicolumn{1}{c}{\our}                           \\
  \cline{5-10} \multicolumn{1}{c}{}   & 
     \multicolumn{1}{c}{Data}   & 
     \multicolumn{1}{c}{TA / HC}   & 
     \multicolumn{1}{c}{$\epsilon$}   & \multicolumn{1}{c}{\#S}  & \multicolumn{1}{c}{T1(\%)}                          & \multicolumn{1}{c|}{T1(\%)}                      & \multicolumn{1}{c}{\#S}   &  \multicolumn{1}{c}{T1(\%)}                          & \multicolumn{1}{c}{T1(\%)}                           \\
 \cline{1-2}
 \cline{3-10} 
SPSA  &V10K&UN  &4 ($L_{\infty}$)  & 4911 &      0                          &     {71}                          &5789       &        0                           &        {58}                                     \\
Boundary &V10K&UN& 2 ($L_{2}$)    & 10000 &     0                          &   {50}                                  & 10000 &    0                &         {50}                                                             \\
Spatial  &V10K&UN  & 2 ($L_{2}$) & 2624 &      0                          &     {36}                          &2634       &        0                           &        {30}                                     \\
\hline
\end{tabular}%
\caption{\textbf{Performance on Gradient-Free Attacks.} Top-1 ( \%) classification accuracy comparison between SoftMax (SMax) and \our. Legends are same as in Tab. \ref{tbl:grad-based}. }
\label{tbl:grad-free}
\end{table}

The consistent superior performance on both gradient-based (both restricted and unrestricted) and gradient free attack shows \our~is not masking the gradients and is attack method agnostic.

\section{Analysis}\label{sec:ablation}

\subsection{Accuracy vs number of layers ($k$)}
\begin{figure}[htb]
    \centering
    \includegraphics[width=0.9\linewidth]{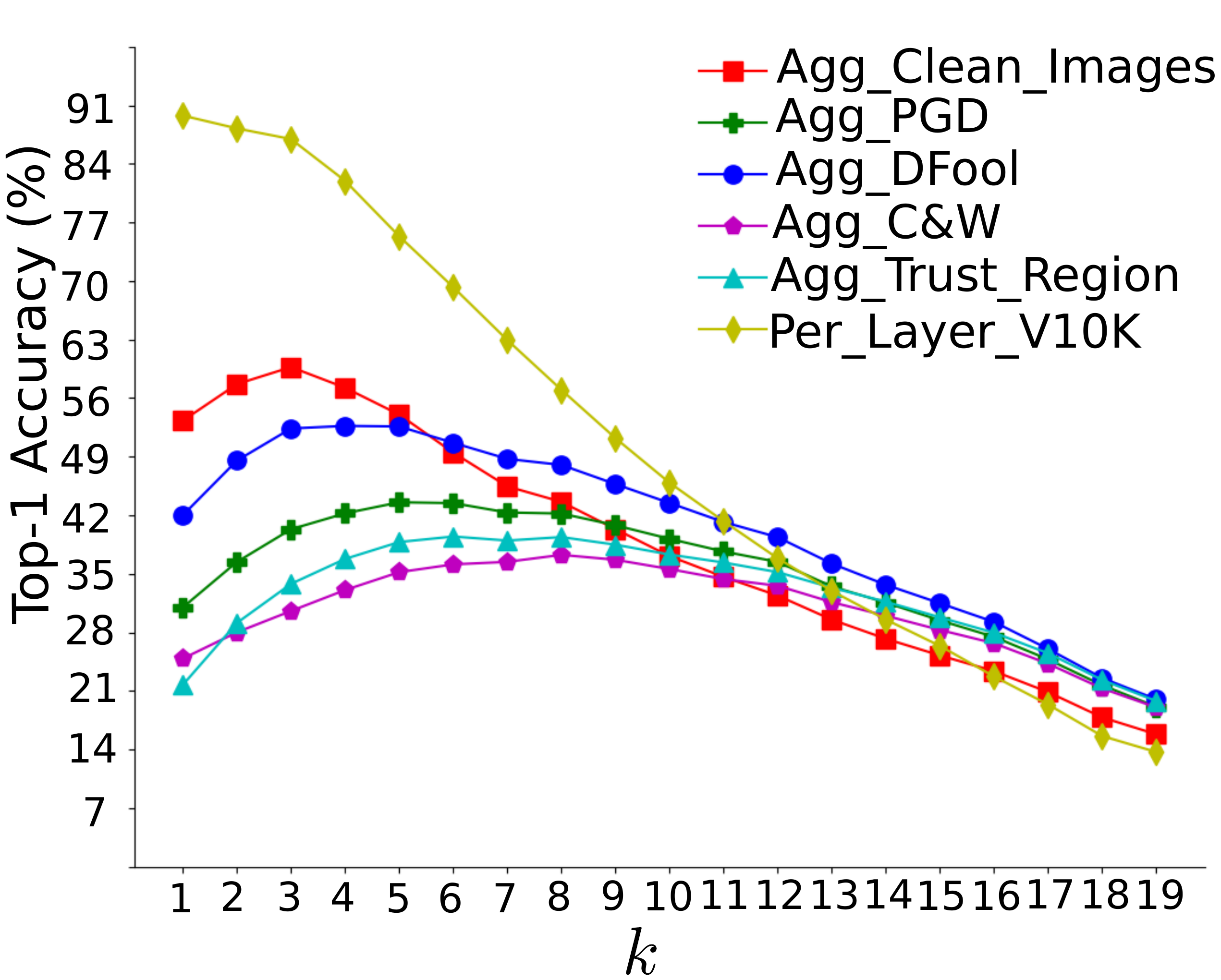}
    \caption{Accuracy vs no. of layers ($k$)  }
    \label{fig:acc_vs_k}
\end{figure}

We report performance of \our~on various attacks reported in Tab. \ref{tbl:grad-based} for all possible values of $k$. The accuracy of VGG-19 w.r.t. the various values of $k$ is plotted in Fig. \ref{fig:acc_vs_k}. We observe a similar accuracy vs $k$ graph for ResNet-50 and note that a reasonable choice of $k$ made based on this graph does not significantly impact \our's performance. Refer Fig. \ref{fig:acc_vs_k}, the `Agg' stands for using aggregated Borda count $B^{:ky}$. PGD(V10K,UN), DFool, C\&W(V10K,UN) and Trust Region are the same experiments as reported in Tab. \ref{tbl:grad-based}, but with all possible values of $k$. `Per\_Layer\_V10K' stands for evaluation using per layer Borda count i.e  \plbl$= \textit{argmax}_{y} ~~B^{\ell y}$ on a separate 10,000 correctly classified subset of validation set. In all our experiments we choose the $k$-highest layers where `Per\_Layer\_V10K' has at-least $75\%$ accuracy. A reasonable change in this accuracy criteria of $75\%$  would not affect the results on adversarial attacks significantly. However, a substantial change (to say $50\%$) deteriorates the performance on clean sample significantly. 
\begin{figure*}[htb]
    \centering
    \includegraphics[width=1.0\linewidth]{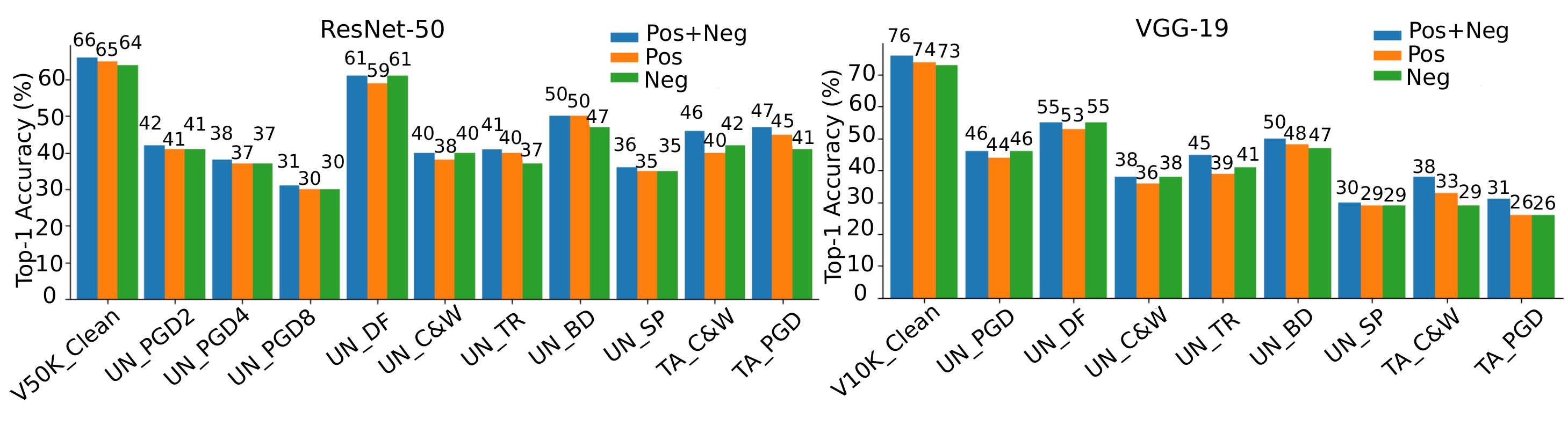}
    \caption{ Effect of Considering Positive and Negative Pre-Activation Responses}
    \label{fig:ablation}
\end{figure*}
The phenomenon of decrease in accuracy of clean samples vs robustness has been studied in \cite{dohmatob2018limitations} and \cite{tsipras2018robustness}. 

\subsection{Effect of positive and negative pre-activation responses}
We report the impact of using positive, negative and a combination of both pre-activation responses on the performance of \our~ in Fig. \ref{fig:ablation}. We consider three variants of Borda count rank aggregation from later $k$ layers. {Pos:} $B^{:ky} = \sum_{\ell=n-k+1}^{n} B^{\ell y}_{+}$, {Neg:} $B^{:ky} = \sum_{\ell=n-k+1}^{n} B^{\ell y}_{-}$, and { Pos+Neg:} $B^{:ky} = \sum_{\ell=n-k+1}^{n} B^{\ell y}_{+}+B^{\ell y}_{-}$. We report the Top-1 accuracy (\%) of the attacks experiment as set up in Tab. \ref{tbl:grad-based} (DF: DFool, C\&W, TR: Trust Region), in Tab. \ref{tbl:grad-free} (BD: Boundary, SP: Spatial), and in Fig. \ref{fig:pgd_strength} (PGD2, PGD4 and PGD8, with $\epsilon=2$, $4$ and $8$ respectively). From the bar chart it is evident that in some experiments, {Pos} performs better than { Neg } (e.g UN\_TR), while in others {Neg} is better than {Pos} only (e.g UN\_DF). It is also evident that {Pos}+{Neg} occasionally improve the overall performance, and the improvement seems significant in the targeted C\&W attacks for both the ResNet-50 and VGG-19. We leave it to the design choice of the application, if inference time is an important parameter, then one may choose either Pos or Neg to reduce the inference time to approximately half of what is reported in Tab. \ref{tbl:inf_time}.

\subsection{Results on CIFAR-10}
While we mainly show results on large-scale dataset (ImageNet), we believe scaling down the datasets to one like CIFAR10 will not have a substantial impact on REGroup’s performance. We evaluate REGroup on CIFAR10 dataset using VGG-19 based classifier. We construct generative classifiers using CIFAR-10 dataset following the same protocol as for the ImageNet case described in the Sec. \ref{sec:LGC}. We apply PGD attack with $\epsilon=4$ and generate adversarial examples. The results are included in the Tab. \ref{tbl:cifar_pgd}.

\begin{table}[!htb]
\centering
\setlength{\tabcolsep}{3pt}
\scriptsize
\begin{tabular}{lccc}
\hline
                  \multicolumn{1}{c}{}                                                                                                                        & \multicolumn{3}{c}{VGG-19}                                                                                                                          \\

 \cline{2-4}                       & \multicolumn{1}{c}{}   &  \multicolumn{1}{c}{SMax}                          & \multicolumn{1}{c}{\our}                           \\
  
     \multicolumn{1}{c}{}                       & \multicolumn{1}{c}{\#S}   &  \multicolumn{1}{c}{T1(\%)}                          & \multicolumn{1}{c}{T1(\%)}                           \\
 \cline{1-2}
 \cline{3-4} 

 \vspace{0.01cm}
Clean  &  10000       &        92                   &   88                                                                      \\
PGD Untargeted  &  9243       &        0                   &   57                                                                      \\
\hline
\end{tabular}%
\caption{\textbf{Performance on CIFAR10.} Comparison of Top-1 classification accuracy between SoftMax (SMax) and \our~ based final classification. $\#S$ is the number of images for which the attacker is successfully able to generate adversarial examples using PGD attack and the accuracies are reported with respect to the $\#S$ samples, hence the 0\% accuracies with the SoftMax (SMax). }
\label{tbl:cifar_pgd}
\end{table}

\subsection{Inference time using \our} Since we suggest to use \our~as a test time replacement of SoftMax, we compare the inference time on both CPU and GPU in Tab. \ref{tbl:inf_time}. We use a machine with an i7-8700 CPU and GTX 1080 GPU.

\begin{table}[!htb]
\centering
\scriptsize
\begin{tabular}{lcccc|cccc}
\hline
                  \multicolumn{1}{c}{}   &  \multicolumn{4}{c|}{ResNet-50}                                                                                                                       & \multicolumn{4}{c}{VGG-19}                                                                                                                          \\

 \cline{2-9}&  \multicolumn{2}{c}{SMax}&   \multicolumn{2}{c|}{\our}        & \multicolumn{2}{c}{SMax}& \multicolumn{2}{c}{\our}                          \\
 \cline{2-9}   \multicolumn{1}{c}{}   &  \multicolumn{1}{c}{GPU}  & \multicolumn{1}{c}{CPU}& \multicolumn{1}{c}{GPU}                          & \multicolumn{1}{c|}{CPU}                     & \multicolumn{1}{c}{GPU}   &  \multicolumn{1}{c}{CPU}
 &  \multicolumn{1}{c}{GPU}& \multicolumn{1}{c}{CPU}                           \\
 \cline{1-2}
 \cline{3-9} 

 \vspace{0.01cm}
Time(s) & 0.02    &  0.06    &0.13                     &    0.35                                         &0.03         &        0.12                   &       0.16    &  0.64                                                              \\
\hline
\end{tabular}%
\caption{\textbf{Inference time comparison.} \our~vs SoftMax}
\label{tbl:inf_time}
\end{table}

In this work, we have presented a simple, scalable, and practical  defense strategy that is model agnostic and does not require any re-training or fine-tuning. We suggest to use \our~at test time to make a pre-trained network robust to adversarial perturbations. There are three aspects of \our~that justify its success. First, instead of using a maximum likelihood based prediction, \our~adopts a ranking preference based approach. Second, aggregation of preferences from multiple layers lead to group decision making, unlike SoftMax that relies on the output of the last layer only. Using both positive and negative layerwise responses help contribute to the robustness of \our. Third, there exists inherent robustness of Borda count based rank aggregation in the presence of noisy individual voters \cite{rothe2019borda},~\cite{kahng2019statistical}. Hence, where SoftMax fails to predict the correct class of an adversarial example, \our~ takes ranked preferences from multiple layers and builds a consensus using Borda count to make robust predictions. Our promising empirical results indicate that deeper theoretical analysis of \our~ would be an interesting direction to pursue.  One direction of analysis could be inspired from the recently proposed perspective of neurons as cooperating classifiers \cite{davel2020dnns}. 

%% file: sections/supp.tex
\section{Hyper-parameters for Generating Adversarial Examples}
We use Foolbox's \cite{rauber2017foolbox} implementation of almost all the adversarial attacks(except SPSA\footnote{https://github.com/tensorflow/cleverhans}, Trust Region\footnote{https://github.com/amirgholami/TRAttack} and cAdv\footnote{https://github.com/AI-secure/Big-but-Invisible-Adversarial-Attack}) used in this work. We report the attack specific hyper-parameters in Tab.\ref{tab:hyperparams}.

\section{Elastic-Net Attacks}
We evaluate \our~ on Elastic-Net attacks \cite{chen2018ead}. Elastic-Net attack formulate the attack process as a elastic-net regularized optimization problem. The results are shown in the table \ref{tbl:ead_attacks}.

\begin{table}[!htb]
\centering
\setlength{\tabcolsep}{3pt}
\scriptsize
\begin{tabular}{l|c|cc|c|cc}
\hline
                  \multicolumn{1}{c}{}   &  \multicolumn{3}{c|}{ResNet-50}                                                                                                                       & \multicolumn{3}{c}{VGG-19}                                                                                                                          \\

 \cline{2-7}   \multicolumn{1}{c|}{}   &  \multicolumn{1}{c|}{}  & \multicolumn{1}{c}{SMax}                          & \multicolumn{1}{c|}{\our}                      & \multicolumn{1}{c|}{}   &  \multicolumn{1}{c}{SMax}                          & \multicolumn{1}{c}{\our}                           \\
  
     \multicolumn{1}{c|}{Attacks}   &  \multicolumn{1}{c|}{\#S}  & \multicolumn{1}{c}{T1(\%)}                          & \multicolumn{1}{c|}{T1(\%)}                      & \multicolumn{1}{c|}{\#S}   &  \multicolumn{1}{c}{T1(\%)}                          & \multicolumn{1}{c}{T1(\%)}                           \\
 \cline{1-2}
 \cline{3-7} 

 \vspace{0.01cm}
EAD-Attack  &  2000   &  0                         &   52                                            &  2000       &        0                   &   49                                                                      \\
\hline
\end{tabular}%
\caption{\textit{Performance on EAD attacks.} Top-1 ( \%) classification accuracy comparison between SoftMax (SMax) and \our. $\#S$ is the number of images for which the attacker is successfully able to generate adversarial examples and the accuracies are reported with respect to the $\#S$ samples, hence the 0\% accuracies with the SoftMax (SMax).   }
\label{tbl:ead_attacks}
\end{table}

\section{Accuracy vs no. of layer/voters(ResNet50)}
\label{sec:resnet50_vs_k}
\begin{figure*}[!htb]
    \centering
    \includegraphics[width=17cm]{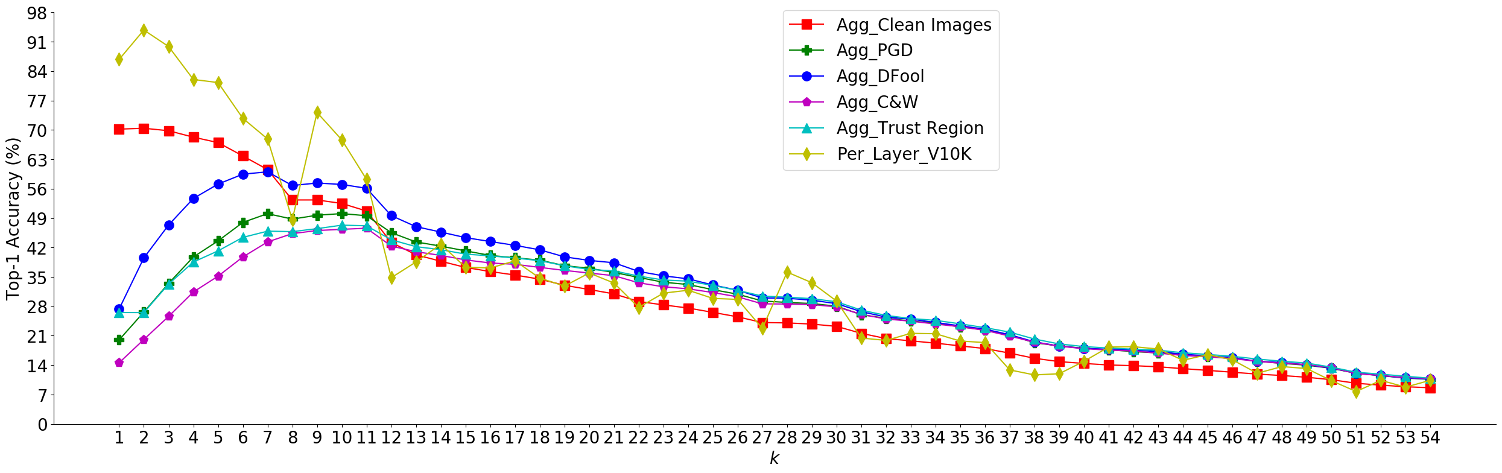}
    \caption{\text{Ablation study for accuracy vs no. of layers ($k$) on ResNet-50}: `Agg' stands for using aggregated Borda count $B^{:ky}$. PGD, DFool, C\&W and Trust Region are the same experiments as reported in table 2 of the main paper, but with all possible values of $k$. "Per\_Layer\_V10K" stands for evaluation using per layer Borda count i.e  \plbl$= \textit{argmax}_{y} ~~B^{\ell y}$ on a separate 10,000 correctly classified subset of validation set. In all our experiments we choose the $k$-highest layers where `Per\_Layer\_V10K' has at-least $75\%$ accuracy. A reasonable change in this accuracy criteria of $75\%$  would not affect the results on adversarial attacks significantly. However, a substantial change (to say $50\%$) deteriorates the performance on clean sample significantly. The phenomenon of decrease in accuracy of clean samples vs robustness has been studied in \cite{dohmatob2018limitations} and \cite{tsipras2018robustness}. \textbf{Note:} There are four down-sampling layers in the ResNet-50 architecture, hence the total 54 layers.   }
    \label{fig:abl_vs_k_resnet}
\end{figure*}

We report the performance of REGroup on various attacks reported in table 2 of the main paper for all possible values of $k$. The accuracy of ResNet-50 w.r.t. the various values of $k$ is plotted in figure \ref{fig:abl_vs_k_resnet}.

\begin{table*}[!htb]
    \centering
    \begin{tabular}{|c|c|}
    \hline
    \textbf{Attack} & \textbf{Hyper-parameters} \\
    \hline
    \multirow{2}{*}{PGD (Untargeted)} & $\epsilon=4$, Dist:$L_\infty$, random\_start=True,\\
    & stepsize=0.01, max\_iter=40  \\
    \hline
    \multirow{2}{*}{DeepFool (Untargeted)} & $\epsilon=2$, Dist:$L_2$, max\_iter=100, \\
    & subsample=10 (Limit on the number of the most likely classes)\\
    \hline
    \multirow{2}{*}{CW (Untargeted)} & $\epsilon=4$, Dist:$L_2$, binary\_search\_steps=5, max\_iter=1000,\\
    &  confidence=0, learning\_rate=0.005, initial\_const=0.01 \\
    \hline
    Trust Region (Untargeted) & $\epsilon=2$, Dist:$L_\infty$, iterations=5000  \\
    \hline
    \multirow{7}{*}{Boundary (Untargeted)} & $\epsilon=2$, Dist:$L_2$, iterations=500, max\_directions=25,  starting\_point=None, \\
    &initialization\_attack=None, 
     log\_every\_n\_steps=None, \\
    & spherical\_step=0.01, source\_step=0.01, step\_adaptation=1.5, \\
    & batch\_size=1, tune\_batch\_size=True, \\
    & threaded\_rnd=True, threaded\_gen=True \\
    \hline
    \multirow{3}{*}{Spatial (Untargeted)} & $\epsilon=2$, Dist=$L_2$, do\_rotations=True, do\_translations=True, x\_shift\_limits=(-5, 5),\\
    & y\_shift\_limits=(-5, 5), angular\_limits=(-5, 5), granularity=10,\\ & random\_sampling=False, abort\_early=True \\
    \hline
    \multirow{2}{*}{PGD (Targeted)} &  Dist = $L_\infty$, binary\_search=True, epsilon=0.3, \\
    & stepsize=0.01, iterations=40, random\_start=True, return\_early=True \\
    \hline
     \multirow{2}{*}{CW (Targeted)} & binary\_search\_steps=5, max\_iterations=1000, confidence=0,  \\
    & learning\_rate=0.005, initial\_const=0.01, abort\_early=True \\
    \hline
    SPSA & $\epsilon=(4,8)$, Dist:$L_\infty$, max\_iter=300, batch\_size=64, early\_stop\_loss\_thresh = 0, \\ 
    & perturbation\_size $\delta=0.01$, Adam LR=0.01\\
    \hline
    \multirow{2}{*}{EAD} & Dist=$L_2$, binary\_search\_steps=5, max\_iterations=1000, confidence=0, \\
    & initial\_learning\_rate=0.01, regularization=0.01, initial\_const=0.01, abort\_early=True \\
    \hline
    \multirow{2}{*}{PGD (Untargeted,HC)} & min\_conf=0.9, Dist=$L_\infty$, binary\_search=True, epsilon=0.3, \\
    & stepsize=0.01, iterations=40, random\_start=True, return\_early=True 
    \\
    \hline
    \multirow{2}{*}{PGD (Targeted,HC)} & min\_conf=0.9, Dist=$L_\infty$, binary\_search=True, epsilon=0.3, \\
    & stepsize=0.01, iterations=40, random\_start=True, return\_early=True  \\
    \hline

    \end{tabular}
\caption{Attack Specific Hyper-parameters.} 
\label{tab:hyperparams}
\end{table*}

\begin{figure*}[!htb]
    \centering
   \hspace{-0.5cm} \subfigure[ResNet-50 $46^{th}$ Layer (CONV)]{
    \includegraphics[width=5.5cm]{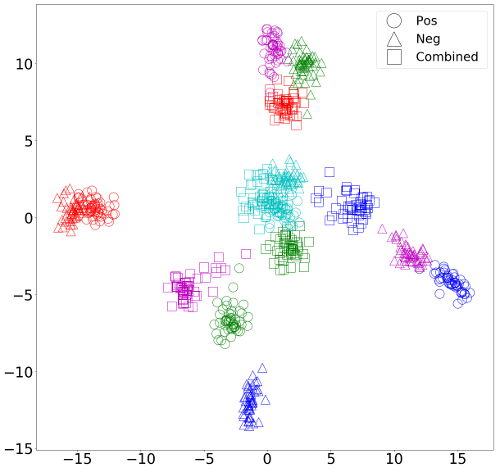}} \hspace{-0.2cm}
     \subfigure[ResNet-50 $48^{th}$ Layer (CONV)]{\includegraphics[width=5.5cm]{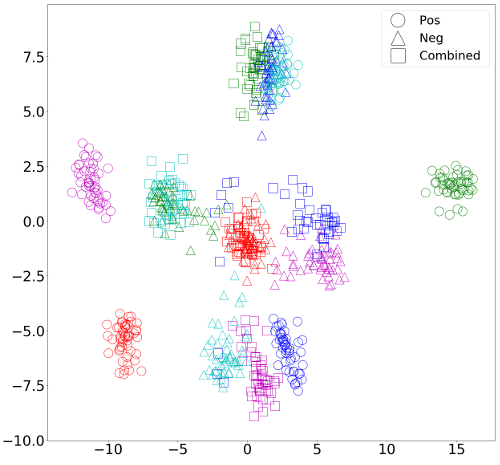}}
    \subfigure[ResNet-50 $50^{th}$ Layer (FC)]{\includegraphics[width=5.5cm]{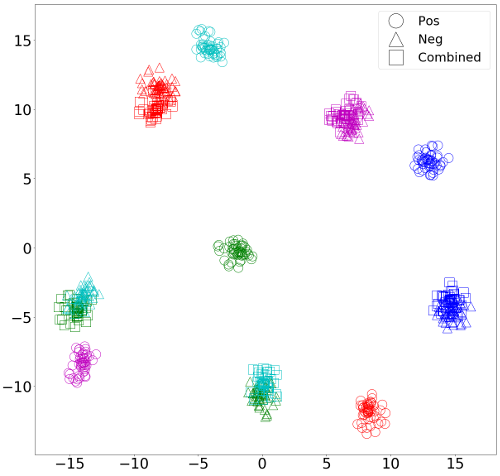}}\\
   \hspace{-0.5cm}     \subfigure[VGG-19 $17^{th}$ Layer (FC)]{\includegraphics[width=5.5cm]{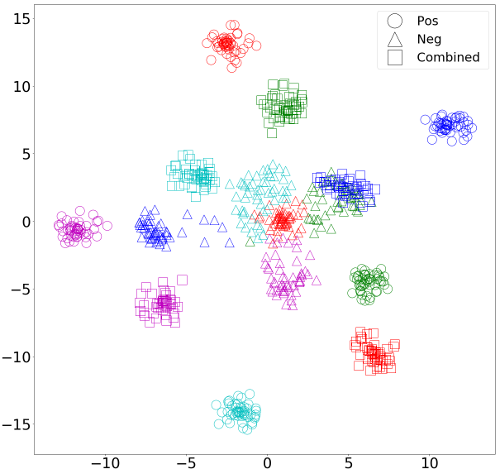}}
   \subfigure[VGG-19 $18^{th}$ Layer (FC)]{\includegraphics[width=5.5cm]{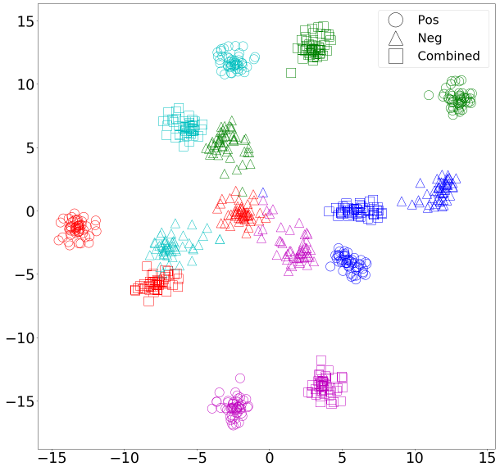}} \hspace{-0.2cm}
     \subfigure[VGG-19 $19^{th}$ Layer (FC)]{\includegraphics[width=5.5cm]{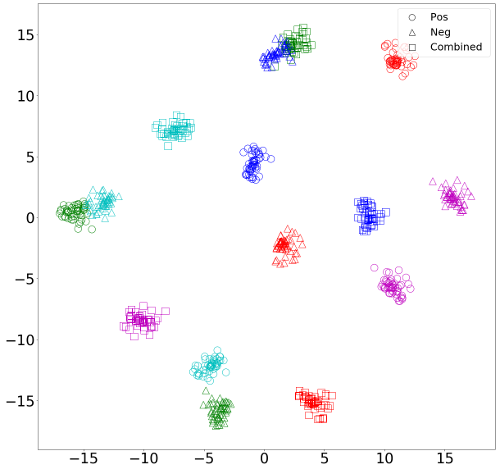}}
    \caption{TSNE visualization of three variants of pre-activation features i.e positive only (pos), negative only (neg) and combined positive and negative (combined). Visualization of 50 samples of 5 random classes of ImageNet dataset. Class membership is color coded. The dimensions of the pos, neg and combined variants of pre-activation feature is the same for any fully connected layer, while for a CONV layer, pos and neg has the same dimension which is equal to the no. of filters/feature maps of the respective CONV layer and for combined it is equal to the dimension we get after flattening the whole CONV layer. It can be observed in figure(b) that the cluster formed by combined pre-activation feature responses is not a tight as formed by pos and neg separately, which shows the importance of considering pos and neg re-activation responses separately.  }
    \label{fig:tsne_plots}
\end{figure*}

\section{Analyzing Pre-Activation Responses}
\label{sec:tsne}
One of the contributions of our proposed approach is to use both positive and negative pre-activation values separately. We observed both positive and negative pre-activation values contain information that can help correctly classify adversarially perturbed samples. An empirical validation of our statement is shown in figure 3 of the main paper. We further show using TSNE~\cite{vanDerMaaten2008} plots that all the three variants of the pre-activation feature of a single layer i.e positive only (pos), negative only (neg) and combined positive and negative pre-activation values forms clusters. This indicates that all three contain equivalent information for discriminating samples from others. While on one hand where ReLU like activation functions discard the negative pre-activation responses, we consider negative responses equivalently important and leverage them to model the layerwise behaviour of class samples. The benefit of using positive and negative accumulators is it reduce the computational cost significantly e.g flattening a convolution layer gives a very high-dimensional vector while accumulator reduce it to number of filter dimensions.

%% file: ms.bbl
\begin{thebibliography}{10}\itemsep=-1pt

\bibitem{akhtar2018defense}
Naveed Akhtar, Jian Liu, and Ajmal Mian.
\newblock Defense against universal adversarial perturbations.
\newblock In {\em CVPR}, pages 3389--3398, 2018.

\bibitem{athalye2018robustness}
Anish Athalye and Nicholas Carlini.
\newblock On the robustness of the cvpr 2018 white-box adversarial example
  defenses.
\newblock {\em arXiv preprint arXiv:1804.03286}, 2018.

\bibitem{athalye2018obfuscated}
Anish Athalye, Nicholas Carlini, and David Wagner.
\newblock Obfuscated gradients give a false sense of security: Circumventing
  defenses to adversarial examples.
\newblock In {\em ICML}, pages 274--283, 2018.

\bibitem{bhagoji2018enhancing}
Arjun~Nitin Bhagoji, Daniel Cullina, Chawin Sitawarin, and Prateek Mittal.
\newblock Enhancing robustness of machine learning systems via data
  transformations.
\newblock In {\em 2018 52nd Annual Conference on Information Sciences and
  Systems (CISS)}, pages 1--5. IEEE, 2018.

\bibitem{bhattad2020unrestricted}
Anand Bhattad, Min~Jin Chong, Kaizhao Liang, Bo Li, and DA Forsyth.
\newblock Unrestricted adversarial examples via semantic manipulation.
\newblock In {\em ICLR}, 2020.

\bibitem{black1958theory}
Duncan Black et~al.
\newblock The theory of committees and elections.
\newblock 1958.

\bibitem{brendel2017decision}
Wieland Brendel, Jonas Rauber, and Matthias Bethge.
\newblock Decision-based adversarial attacks: Reliable attacks against
  black-box machine learning models.
\newblock {\em arXiv preprint arXiv:1712.04248}, 2017.

\bibitem{carlini2019evaluating}
Nicholas Carlini, Anish Athalye, Nicolas Papernot, Wieland Brendel, Jonas
  Rauber, Dimitris Tsipras, Ian Goodfellow, and Aleksander Madry.
\newblock On evaluating adversarial robustness.
\newblock {\em arXiv preprint arXiv:1902.06705}, 2019.

\bibitem{carlini2017adversarial}
Nicholas Carlini and David Wagner.
\newblock Adversarial examples are not easily detected: Bypassing ten detection
  methods.
\newblock In {\em Proceedings of the 10th ACM Workshop on Artificial
  Intelligence and Security}, pages 3--14. ACM, 2017.

\bibitem{carlini2017towards}
Nicholas Carlini and David Wagner.
\newblock Towards evaluating the robustness of neural networks.
\newblock In {\em 2017 IEEE Symposium on Security and Privacy (SP)}, pages
  39--57. IEEE, 2017.

\bibitem{chakraborty2018adversarial}
Anirban Chakraborty, Manaar Alam, Vishal Dey, Anupam Chattopadhyay, and Debdeep
  Mukhopadhyay.
\newblock Adversarial attacks and defences: A survey.
\newblock {\em arXiv preprint arXiv:1810.00069}, 2018.

\bibitem{chen2018ead}
Pin-Yu Chen, Yash Sharma, Huan Zhang, Jinfeng Yi, and Cho-Jui Hsieh.
\newblock Ead: elastic-net attacks to deep neural networks via adversarial
  examples.
\newblock In {\em AAAI}, 2018.

\bibitem{chin2011accelerated}
Tat-Jun Chin, Jin Yu, and David Suter.
\newblock Accelerated hypothesis generation for multistructure data via
  preference analysis.
\newblock {\em IEEE TPAMI}, 34(4):625--638, 2011.

\bibitem{davel2020dnns}
Marelie Davel, Marthinus Theunissen, Arnold Pretorius, and Etienne Barnard.
\newblock Dnns as layers of cooperating classifiers.
\newblock {\em Proceedings of the AAAI Conference on Artificial Intelligence},
  34(04):3725–3732, Apr 2020.

\bibitem{imagenet}
J. Deng, W. Dong, R. Socher, L.-J. Li, K. Li, and L. Fei-Fei.
\newblock {ImageNet: A Large-Scale Hierarchical Image Database}.
\newblock In {\em CVPR09}, 2009.

\bibitem{dohmatob2018limitations}
Elvis Dohmatob.
\newblock Limitations of adversarial robustness: strong no free lunch theorem.
\newblock {\em arXiv preprint arXiv:1810.04065}, 2018.

\bibitem{engstrom2018evaluating}
Logan Engstrom, Andrew Ilyas, and Anish Athalye.
\newblock Evaluating and understanding the robustness of adversarial logit
  pairing.
\newblock {\em arXiv preprint arXiv:1807.10272}, 2018.

\bibitem{engstrom2019exploring}
Logan Engstrom, Brandon Tran, Dimitris Tsipras, Ludwig Schmidt, and Aleksander
  Madry.
\newblock Exploring the landscape of spatial robustness.
\newblock In {\em ICML}, pages 1802--1811, 2019.

\bibitem{Fetaya_ICLR2020}
Ethan Fetaya, Joern-Henrik Jacobsen, Will Grathwohl, and Richard Zemel.
\newblock Understanding the limitations of conditional generative models.
\newblock In {\em ICLR}, 2020.

\bibitem{grosse2017statistical}
Kathrin Grosse, Praveen Manoharan, Nicolas Papernot, Michael Backes, and
  Patrick McDaniel.
\newblock On the (statistical) detection of adversarial examples.
\newblock {\em arXiv preprint arXiv:1702.06280}, 2017.

\bibitem{guo2017countering}
Chuan Guo, Mayank Rana, Moustapha Cisse, and Laurens Van Der~Maaten.
\newblock Countering adversarial images using input transformations.
\newblock {\em arXiv preprint arXiv:1711.00117}, 2017.

\bibitem{he2015delving}
Kaiming He, Xiangyu Zhang, Shaoqing Ren, and Jian Sun.
\newblock Delving deep into rectifiers: Surpassing human-level performance on
  imagenet classification.
\newblock In {\em ICCV}, pages 1026--1034, 2015.

\bibitem{hinton2012deep}
Geoffrey Hinton, Li Deng, Dong Yu, George Dahl, Abdel-rahman Mohamed, Navdeep
  Jaitly, Andrew Senior, Vincent Vanhoucke, Patrick Nguyen, Brian Kingsbury,
  et~al.
\newblock Deep neural networks for acoustic modeling in speech recognition.
\newblock {\em IEEE Signal processing magazine}, 29, 2012.

\bibitem{Chin_NIPS2009}
Tat jun Chin, Hanzi Wang, and David Suter.
\newblock The ordered residual kernel for robust motion subspace clustering.
\newblock In Y. Bengio, D. Schuurmans, J.~D. Lafferty, C.~K.~I. Williams, and
  A. Culotta, editors, {\em NIPS}, pages 333--341. Curran Associates, Inc.,
  2009.

\bibitem{kahng2019statistical}
Anson Kahng, Min~Kyung Lee, Ritesh Noothigattu, Ariel Procaccia, and
  Christos-Alexandros Psomas.
\newblock Statistical foundations of virtual democracy.
\newblock In {\em ICML}, pages 3173--3182, 2019.

\bibitem{kannan2018adversarial}
Harini Kannan, Alexey Kurakin, and Ian Goodfellow.
\newblock Adversarial logit pairing.
\newblock {\em arXiv preprint arXiv:1803.06373}, 2018.

\bibitem{karpathy2014large}
Andrej Karpathy, George Toderici, Sanketh Shetty, Thomas Leung, Rahul
  Sukthankar, and Li Fei-Fei.
\newblock Large-scale video classification with convolutional neural networks.
\newblock In {\em CVPR}, pages 1725--1732, 2014.

\bibitem{krizhevsky2012imagenet}
Alex Krizhevsky, Ilya Sutskever, and Geoffrey~E Hinton.
\newblock Imagenet classification with deep convolutional neural networks.
\newblock In {\em NIPS}, pages 1097--1105, 2012.

\bibitem{kurakin2016adversarial}
Alexey Kurakin, Ian Goodfellow, and Samy Bengio.
\newblock Adversarial machine learning at scale.
\newblock {\em arXiv preprint arXiv:1611.01236}, 2016.

\bibitem{li2017adversarial}
Xin Li and Fuxin Li.
\newblock Adversarial examples detection in deep networks with convolutional
  filter statistics.
\newblock In {\em ICCV}, pages 5764--5772, 2017.

\bibitem{liao2018defense}
Fangzhou Liao, Ming Liang, Yinpeng Dong, Tianyu Pang, Xiaolin Hu, and Jun Zhu.
\newblock Defense against adversarial attacks using high-level representation
  guided denoiser.
\newblock In {\em CVPR}, pages 1778--1787, 2018.

\bibitem{robustml}
Aleksander Madry, Athalye Anish, Tsipras Dimitris, and Engstrom Logan.
\newblock https://www.robust-ml.org/.
\newblock \url{https://www.robust-ml.org/}, 2020.
\newblock Accessed: 2020-01-05.

\bibitem{madry2017towards}
Aleksander Madry, Aleksandar Makelov, Ludwig Schmidt, Dimitris Tsipras, and
  Adrian Vladu.
\newblock Towards deep learning models resistant to adversarial attacks.
\newblock {\em arXiv preprint arXiv:1706.06083}, 2017.

\bibitem{metzen2017detecting}
Jan~Hendrik Metzen, Tim Genewein, Volker Fischer, and Bastian Bischoff.
\newblock On detecting adversarial perturbations.
\newblock {\em arXiv preprint arXiv:1702.04267}, 2017.

\bibitem{moosavi2017universal}
Seyed-Mohsen Moosavi-Dezfooli, Alhussein Fawzi, Omar Fawzi, and Pascal
  Frossard.
\newblock Universal adversarial perturbations.
\newblock In {\em CVPR}, pages 1765--1773, 2017.

\bibitem{moosavi2016deepfool}
Seyed-Mohsen Moosavi-Dezfooli, Alhussein Fawzi, and Pascal Frossard.
\newblock Deepfool: a simple and accurate method to fool deep neural networks.
\newblock In {\em CVPR}, pages 2574--2582, 2016.

\bibitem{pang2019rethinking}
Tianyu Pang, Kun Xu, Yinpeng Dong, Chao Du, Ning Chen, and Jun Zhu.
\newblock Rethinking softmax cross-entropy loss for adversarial robustness.
\newblock In {\em ICLR}, 2020.

\bibitem{papernot2017extending}
Nicolas Papernot and Patrick McDaniel.
\newblock Extending defensive distillation.
\newblock {\em arXiv preprint arXiv:1705.05264}, 2017.

\bibitem{papernot2017practical}
Nicolas Papernot, Patrick McDaniel, Ian Goodfellow, Somesh Jha, Z~Berkay Celik,
  and Ananthram Swami.
\newblock Practical black-box attacks against machine learning.
\newblock In {\em Proceedings of the 2017 ACM on Asia conference on computer
  and communications security}, pages 506--519, 2017.

\bibitem{papernot2016distillation}
Nicolas Papernot, Patrick McDaniel, Xi Wu, Somesh Jha, and Ananthram Swami.
\newblock Distillation as a defense to adversarial perturbations against deep
  neural networks.
\newblock In {\em 2016 IEEE Symposium on Security and Privacy (SP)}, pages
  582--597. IEEE, 2016.

\bibitem{prakash2018deflecting}
Aaditya Prakash, Nick Moran, Solomon Garber, Antonella DiLillo, and James
  Storer.
\newblock Deflecting adversarial attacks with pixel deflection.
\newblock In {\em CVPR}, pages 8571--8580, 2018.

\bibitem{raff2019barrage}
Edward Raff, Jared Sylvester, Steven Forsyth, and Mark McLean.
\newblock Barrage of random transforms for adversarially robust defense.
\newblock In {\em CVPR}, pages 6528--6537, 2019.

\bibitem{rauber2017foolbox}
Jonas Rauber, Wieland Brendel, and Matthias Bethge.
\newblock Foolbox: A python toolbox to benchmark the robustness of machine
  learning models.
\newblock {\em arXiv preprint arXiv:1707.04131}, 2017.

\bibitem{rothe2019borda}
J{\"o}rg Rothe.
\newblock Borda count in collective decision making: A summary of recent
  results.
\newblock In {\em AAAI}, volume~33, pages 9830--9836, 2019.

\bibitem{samangouei2018defense}
Pouya Samangouei, Maya Kabkab, and Rama Chellappa.
\newblock Defense-gan: Protecting classifiers against adversarial attacks using
  generative models.
\newblock {\em arXiv preprint arXiv:1805.06605}, 2018.

\bibitem{shafahi2020universal}
Ali Shafahi, Mahyar Najibi, Zheng Xu, John~P Dickerson, Larry~S Davis, and Tom
  Goldstein.
\newblock Universal adversarial training.
\newblock In {\em AAAI}, pages 5636--5643, 2020.

\bibitem{sharma2018bypassing}
Yash Sharma and Pin-Yu Chen.
\newblock Bypassing feature squeezing by increasing adversary strength.
\newblock {\em arXiv preprint arXiv:1803.09868}, 2018.

\bibitem{song2018pixeldefend}
Yang Song, Taesup Kim, Sebastian Nowozin, Stefano Ermon, and Nate Kushman.
\newblock Pixeldefend: Leveraging generative models to understand and defend
  against adversarial examples.
\newblock In {\em ICLR}, 2018.

\bibitem{szegedy2013intriguing}
Christian Szegedy, Wojciech Zaremba, Ilya Sutskever, Joan Bruna, Dumitru Erhan,
  Ian Goodfellow, and Rob Fergus.
\newblock Intriguing properties of neural networks.
\newblock {\em arXiv preprint arXiv:1312.6199}, 2013.

\bibitem{tiwari2018dgsac}
Lokender Tiwari and Saket Anand.
\newblock Dgsac: Density guided sampling and consensus.
\newblock In {\em IEEE WACV}, pages 974--982. IEEE, 2018.

\bibitem{tsipras2018robustness}
Dimitris Tsipras, Shibani Santurkar, Logan Engstrom, Alexander Turner, and
  Aleksander Madry.
\newblock Robustness may be at odds with accuracy.
\newblock {\em arXiv preprint arXiv:1805.12152}, 2018.

\bibitem{uesato2018adversarial}
Jonathan Uesato, Brendan O’Donoghue, Pushmeet Kohli, and Aaron Oord.
\newblock Adversarial risk and the dangers of evaluating against weak attacks.
\newblock In {\em ICML}, pages 5025--5034, 2018.

\bibitem{vanDerMaaten2008}
Laurens van~der Maaten and Geoffrey Hinton.
\newblock Visualizing data using {t-SNE}.
\newblock {\em Journal of Machine Learning Research}, 9:2579--2605, 2008.

\bibitem{van1992borda}
Jill Van~Newenhizen.
\newblock The borda method is most likely to respect the condorcet principle.
\newblock {\em Economic Theory}, 2(1):69--83, 1992.

\bibitem{wang2016learning}
Qinglong Wang, Wenbo Guo, Kaixuan Zhang, II Ororbia, G Alexander, Xinyu Xing,
  Xue Liu, and C~Lee Giles.
\newblock Learning adversary-resistant deep neural networks.
\newblock {\em arXiv preprint arXiv:1612.01401}, 2016.

\bibitem{xie2020adversarial}
Cihang Xie, Mingxing Tan, Boqing Gong, Jiang Wang, Alan~L Yuille, and Quoc~V
  Le.
\newblock Adversarial examples improve image recognition.
\newblock In {\em Proceedings of the IEEE/CVF Conference on Computer Vision and
  Pattern Recognition}, pages 819--828, 2020.

\bibitem{xie2017mitigating}
Cihang Xie, Jianyu Wang, Zhishuai Zhang, Zhou Ren, and Alan Yuille.
\newblock Mitigating adversarial effects through randomization.
\newblock {\em arXiv preprint arXiv:1711.01991}, 2017.

\bibitem{xie2017adversarial}
Cihang Xie, Jianyu Wang, Zhishuai Zhang, Yuyin Zhou, Lingxi Xie, and Alan
  Yuille.
\newblock Adversarial examples for semantic segmentation and object detection.
\newblock In {\em ICCV}, pages 1369--1378, 2017.

\bibitem{xie2019feature}
Cihang Xie, Yuxin Wu, Laurens van~der Maaten, Alan~L Yuille, and Kaiming He.
\newblock Feature denoising for improving adversarial robustness.
\newblock In {\em CVPR}, pages 501--509, 2019.

\bibitem{xu2017feature}
Weilin Xu, David Evans, and Yanjun Qi.
\newblock Feature squeezing: Detecting adversarial examples in deep neural
  networks.
\newblock {\em arXiv preprint arXiv:1704.01155}, 2017.

\bibitem{yao2019trust}
Zhewei Yao, Amir Gholami, Peng Xu, Kurt Keutzer, and Michael~W Mahoney.
\newblock Trust region based adversarial attack on neural networks.
\newblock In {\em CVPR}, pages 11350--11359, 2019.

\bibitem{young1988condorcet}
H~Peyton Young.
\newblock Condorcet's theory of voting.
\newblock {\em American Political science review}, 82(4):1231--1244, 1988.

\bibitem{zhang2020clipped}
Zhanyuan Zhang, Benson Yuan, Michael McCoyd, and David Wagner.
\newblock Clipped bagnet: Defending against sticker attacks with clipped
  bag-of-features.
\newblock In {\em 3rd Deep Learning and Security Workshop (DLS)}, 2020.

\bibitem{zheng2016improving}
Stephan Zheng, Yang Song, Thomas Leung, and Ian Goodfellow.
\newblock Improving the robustness of deep neural networks via stability
  training.
\newblock In {\em CVPR}, pages 4480--4488, 2016.

\end{thebibliography}
